\documentstyle[nm2000,epsf]{article}

\newtheorem{definition}{Definition}

\newtheorem{theorem}{Theorem}

\def\and{\mbox{ and }}

\def\imply{\supset}
\def\set{{\:\leftarrow\:\:}}

\def\negg(#1){\overline{#1}}

\def\eql(#1,#2){{#1\!\!=\!\!#2}}

\def\true{{\sf true}}
\def\false{{\sf false}}
\def\low{{\sf low}}
\def\high{{\sf high}}

\def\low{{\it low}}
\def\high{{\it high}}
\def\var{{\it var}}
\def\fbdd2ddnnf{{\sc fbdd2ddnnf}}
\def\pb2dnnf{{\sc 1npb2dnnf}}

\def\atoms{{\it atoms}}
\def\project(#1,#2){{\it project}(#1,#2)}
\def\set{{\leftarrow}}
\def\cache{{\sc cache}}
\def\nil{{\sc nil}}
\def\dnnfs{{\sc cnf2ddnnf}}
\def\bdnnfs{{\sc cl2ddnnf}}

\def\clause{{\it clause}}

\def\mc{{\it Models\#}}

\def\D{{\bf D}}
\def\S{{\bf S}}

\def\ok{{\it ok}}

\def\val{{\sc val}}
\def\pd{{\sc pd}}
\def\cpd{{\sc cpd}}

\def\sel(#1,#2){{(#1,#2)}}
\def\eql(#1,#2){{#1\!\!=#2}}

\begin{document}

\title{On the Tractable Counting of Theory Models and its 
Application to Belief Revision and Truth Maintenance}
\author{Adnan Darwiche \\
Computer Science Department \\
University of California \\
Los Angeles, Ca 90095 \\
{\it darwiche@cs.ucla.edu}}

\maketitle

\begin{abstract}
We introduced decomposable negation normal form (DNNF) recently as a tractable
form of propositional theories, and provided a number of powerful logical operations that can
be performed on it in polynomial time. We also presented an algorithm for compiling 
any conjunctive normal form (CNF) into DNNF and provided a structure-based guarantee on its space and time 
complexity. We present in this paper a linear-time algorithm for converting an ordered binary
decision diagram (OBDD) 
representation of a propositional theory into an equivalent DNNF, showing that DNNFs scale 
as well as OBDDs. We also identify a subclass of DNNF which we call {\em deterministic} DNNF, 
d-DNNF, and show that the previous complexity guarantees on compiling DNNF continue to hold for this stricter 
subclass, which has stronger properties. In particular, we present a new operation on d-DNNF 
which allows us to count its models under the assertion, retraction and flipping of every 
literal by traversing the d-DNNF twice. That is, after such traversal, we can test in 
constant-time: the entailment of any literal by the d-DNNF, and the consistency of the d-DNNF 
under the retraction or flipping of any literal.
We demonstrate the significance of these new operations by showing how they allow us to 
implement linear-time, complete truth maintenance systems and linear-time, complete belief 
revision systems for two important classes of propositional theories.
\end{abstract}

\section{Introduction}

Knowledge compilation has been emerging recently as a new direction of research for dealing with 
the computational intractability of general propositional reasoning \cite{selmanACM,cadoli97}. 
According to this approach, the reasoning process is split into two phases: an off-line compilation 
phase and an on-line query-answering phase. In the off-line phase, the propositional theory is compiled 
into some target language, which is typically a tractable one. In the on-line phase, the compiled target
is used to efficiently answer a (potentially) exponential number of queries. The main motivation
behind knowledge compilation is to push as much of the computational overhead as possible into the off-line
phase, in order to amortize that overhead over all on-line queries.

One of the key aspects of any compilation approach is the target language into which the propositional
theory is compiled. Previous compilation approaches have proposed Horn theories, prime implicates/implicants,
and ordered binary decision diagrams (OBDDs) as targets for such compilation 
\cite{selmanACM,cadoli97,marquis95,boufkhadIJCAI97,Bryant92}. 
A more recent compilation target language is {\em decomposable negation normal form (DNNF)} 
\cite{darwicheIJCAI99a,darwicheJAIR-SSD,darwicheD109}. 
DNNF is universal;
supports a rich set of polynomial-time operations;
is more space--efficient than OBDDs  \cite{darwicheD109};
and is very simplistic as far as its structure and algorithms are concerned.
Propositional theories in DNNF are highly tractable \cite{darwicheD109}:
\begin{enumerate}
\item Deciding the {\em satisfiability} of a DNNF can be done in linear time.

\item {\em Conjoining} a DNNF with a set of literals can be done in linear time.

\item {\em Projecting} a DNNF on some atoms can be done in linear time.
Intuitively, to project a theory on a set of atoms is to compute the strongest sentence entailed
by the theory on these atoms. 

\item Computing the {\em minimum--cardinality} of a DNNF can be done in linear time.
The cardinality of a model is the number of atoms that are set to false in the model.
The minimum--cardinality of a theory is the minimum--cardinality of any of its models.

\item {\em Minimizing} a DNNF can be done in quadratic time.
To minimize a theory is to produce another theory which models are exactly the minimum-cardinality 
models of the original theory.

\item {\em Enumerating} the models of a DNNF can be done in time linear in its size and quadratic
in the number of its models.
\end{enumerate}

This paper rests on two key contributions. First, we show that DNNF representations scale as 
well as OBDD representations \cite{Bryant92} by presenting a linear-time algorithm for converting an OBDD
representation of a propositional theory into an equivalent DNNF representation. 
Second, we identify a subclass of DNNF, 
which we call {\em deterministic DNNF}, d-DNNF, and present a new linear-time operation for counting its 
models under the assertion, retraction and flipping of literals. In particular, we 
show how to traverse a d-DNNF only twice and yet compute: its number of models
under the assertion, retraction and flipping of each literal. This allows us to test in linear time:
the consistency of the d-DNNF under the assertion, retraction and flipping of each literal, therefore,
allowing us to implement linear-time, complete truth maintenance and belief revision systems.

What is interesting though is that two of our key complexity results with respect to DNNF compilations
continue to hold with respect to the subclass of d-DNNF. This includes a structure-based algorithm 
which can compile any CNF into an equivalent DNNF in linear time given that the treewidth of the CNF
is bounded  \cite{darwicheIJCAI99a}. It also includes the newly proposed algorithm for converting 
any OBDD into an equivalent DNNF in linear time.

This paper is structured as follows. We first review DNNF, introduce the class of 
deterministic DNNF and then discuss the new operation for model counting. We follow 
that by discussing the application of this counting operation to truth maintenance,
belief revision, and model-based diagnosis systems.
We finally close with some concluding remarks. 
Proofs of all results are available in the full paper \cite{darwicheD113}.

\section{Deterministic DNNF} \label{sec:d-DNNF}

\begin{figure*}[tb]
\begin{center}
\mbox{\epsfxsize=105mm \leavevmode \epsffile{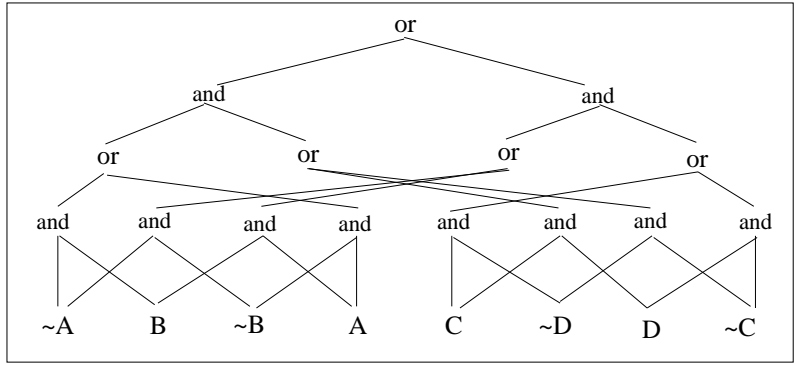}}
\end{center}
\caption{A theory in DNNF. The theory has eight models, 
representing the odd-cardinality models:
\(\neg A \wedge B \wedge C \wedge D\);
\(A \wedge \neg B \wedge C \wedge D\); 
\(A \wedge B \wedge \neg C \wedge D\); 
\(A \wedge B \wedge C \wedge \neg D\); 
\(\neg A \wedge \neg B \wedge \neg C \wedge D\); 
\(\neg A \wedge \neg B \wedge C \wedge \neg D\);
\(\neg A \wedge B \wedge \neg C \wedge \neg D\);
\(A \wedge \neg B \wedge \neg C \wedge \neg D\).
\label{fig:odd}}
\end{figure*}

A propositional sentence is in negation normal form (NNF) if it is constructed from literals using only
the conjoin and disjoin operators.
Figure~\ref{fig:odd} shows a sentence in NNF depicted as a rooted, directed acyclic 
graph where the children of each node are shown below it in the graph. Each leaf node represents
a literal and each non-leaf node represents a conjunction or a disjunction.
We allow \(\true\) and \(\neg \false\) to appear as leaves in a DNNF
to denote a conjunction with no conjuncts. Similarly, we allow \(\false\) and \(\neg \true\) 
as leaves to represent a disjunction with no disjuncts. 
The size of an NNF is measured by the number of edges in its graphical representation. 
Our concern here is mainly with a subclass of NNFs:
\begin{definition}\cite{darwicheIJCAI99a}
A \underline{decomposable negation normal form} (DNNF) is a negation normal form satisfying
decomposability property: for any conjunction \(\bigwedge_i \alpha_i\) appearing in the form,
no atom is shared by any pair of conjuncts in \(\bigwedge_i \alpha_i\).
\end{definition}
The NNF \((A \vee B) \wedge (\neg A \vee C)\) is not decomposable since atom \(A\) is shared
by the two conjuncts. 
But the NNF in Figure~\ref{fig:odd} is decomposable. It has ten conjunctions and the conjuncts
of each share no atoms. Decomposability is the property which makes DNNF tractable.

One of the key operations on DNNF is that of conditioning:
\begin{definition}\cite{darwicheIJCAI99a}
Let \(\alpha\) be a propositional sentence and let \(\gamma\) be an instantiation.\footnote{An 
instantiation of a set of atoms is a conjunction of literals, one literal for each atom in the set.}
The \underline{conditioning} of \(\alpha\) on \(\gamma\), written \(\alpha \mid \gamma\), is the sentence 
resulting from replacing each atom \(p\) in \(\alpha\) with \(\true\) if the +ve literal \(p\) 
appears in \(\gamma\) and with \(\false\) if the -ve literal \(\neg p\) appears in \(\gamma\).
\end{definition}
For example, conditioning the DNNF \((\neg A \wedge \neg B) \vee (B \wedge C)\) on 
instantiation \(B\wedge D\) gives \((\neg A \wedge \neg \true) \vee (\true \wedge C)\) and 
conditioning it on \(\neg B\wedge D\) gives \((\neg A \wedge \neg \false) \vee (\false \wedge C)\).
Conditioning is a key operation because it allows us to conjoin a DNNF \(\Delta\) with some 
instantiation \(\alpha\) (which may share atoms with \(\Delta\)) while ensuring that the result 
is also a DNNF. Specifically, \((\Delta \mid \alpha) \wedge \alpha\) is a DNNF which is equivalent 
to \(\Delta\wedge\alpha\) and can be computed in time linear in the size of \(\Delta\).

We now introduce the class of deterministic DNNF:
\begin{definition}
A \underline{deterministic DNNF (d-DNNF)} is a DNNF satisfying the following property:
for any disjunction \(\bigvee_i \alpha_i\) appearing in the form, every pair of disjuncts 
in \(\bigvee_i \alpha_i\) are disjoint.
\end{definition}
For example, \((A \wedge B) \vee C\) is a DNNF but is not deterministic since the disjuncts
\(A \wedge B\) and \(C\) are not disjoint. However, the DNNF 
\((A \wedge B) \vee (\neg A\wedge C)\) is deterministic. Note that although every DNF is a 
DNNF, not every DNF is a d-DNNF.

\begin{figure}[tb]
\begin{center}
\mbox{\epsfxsize=70mm \leavevmode \epsffile{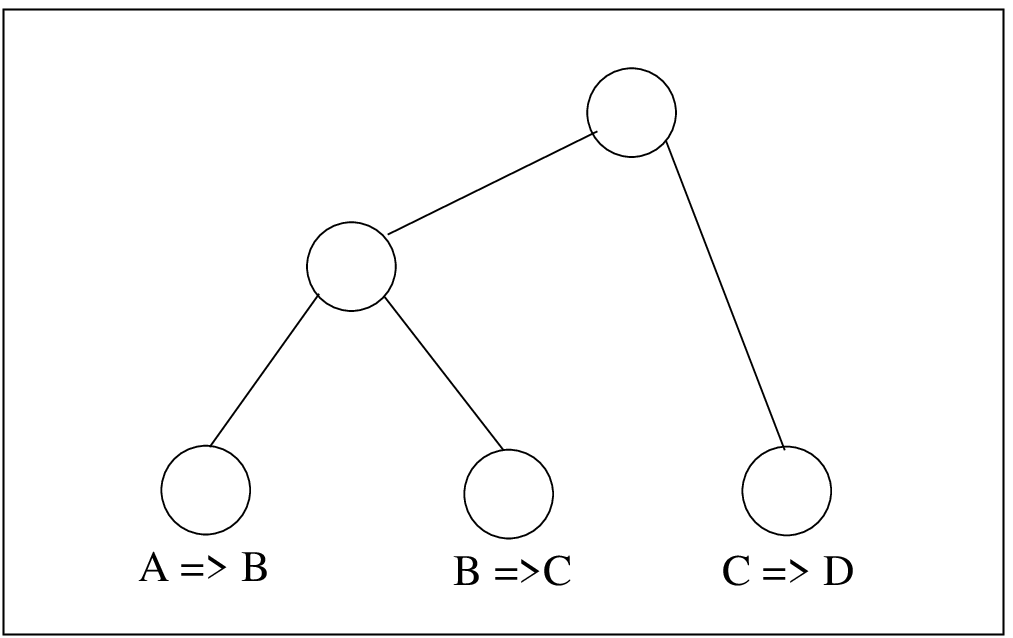}}
\end{center}
\caption{A decomposition tree for the theory
\(A \imply B\), \(B \imply C\), \(C \imply D\). \label{fig:dtree}}
\end{figure}

The main value of d-DNNF is the ability to count its models in linear time and under the assertion,
retraction and flipping of literals. Such operations will be discussed in the following section. 
In the remainder of this
section we present two results that hold for DNNF, but continue to hold for d-DNNF. The first is a structure-based
algorithm that we introduced in \cite{darwicheIJCAI99a} for converting a CNF into a DNNF. The algorithm utilizes
a {\em decomposition tree,}
which is a binary tree the leaves of which correspond to the CNF clauses---see Figure~\ref{fig:dtree}.
Each decomposition tree has a width and the complexity of the algorithm is exponential only in the width of used tree. 
The algorithm is given with a slight modification in Figure~\ref{fig:cnf2ddnnf}.
The only difference between this version and the one in \cite{darwicheIJCAI99a} is that we have 
\(\bdnnfs(\clause(N)\mid\alpha)\) instead of \(\clause(N)\mid\alpha\), 
therefore, converting a clause to a d-DNNF at the boundary condition. 

\begin{theorem}  \label{theo:dDNNF} \cite{darwicheIJCAI99a}
Let \(N\) be the root of decomposition tree \(T\) used in Figure~\ref{fig:cnf2ddnnf}. Then 
\(\dnnfs(N,\alpha)\) will return \(\Delta \mid \alpha\) in DNNF, where \(\Delta\) 
contains the clauses attached to the leaves of \(T\). Moreover, 
the time and space complexity of the algorithm  is \(O(n w 2^w)\), where \(n\) is 
the number of clauses in \(\Delta\) and \(w\) is the width of decomposition tree \(T\).
\end{theorem}
\begin{theorem}
The DNNF returned by the algorithm in  Figure~\ref{fig:cnf2ddnnf} is deterministic.
\end{theorem}
The class of CNF theories with bounded treewidth is defined in \cite{darwicheIJCAI99a}, where it is
shown that, for this class of theories, one can construct in linear time a decomposition tree of 
bounded width. Therefore, one can compile a d-DNNF of linear size for this class of theories.

\begin{figure}[tb]
\begin{center}
\fbox{
\begin{minipage}{3.0in}
%{\small
\begin{center}
\underline{{\bf Algorithm~cnf2ddnnf}}
\end{center}
/* \(N\) is a node in a decomposition tree */ \\
/* \(\alpha\) is an instantiation */
\begin{tabbing}
{\sc c}\={\sc nf2ddnnf}\((N,\alpha)\) \\
 \>\(\psi \set \project(\alpha,{\atoms(N)})\) \\
 \>i\=f \(\cache_N(\psi) \neq \nil\), return \(\cache_N(\psi)\) \\
 \>i\=f \(N\) is a leaf node, \\
 \>\>then \(\gamma \set\) \(\bdnnfs(\clause(N)\mid\alpha)\) \\
 \>\>else \(\gamma\)\=\(\set\bigvee_\beta\)\=\(\dnnfs(N_l,\alpha\wedge\beta)\wedge\) \\
 \>\>\>\>\(\dnnfs(N_r,\alpha\wedge\beta)\wedge\beta\) \\
 \>\> \> where \(\beta\) ranges over all instantiations \\
 \>\> \> of \(\atoms(N_l)\cap\atoms(N_r)-\atoms(\alpha)\) \\
 \>\(\cache_N(\psi) \set \gamma\) \\
 \>return \(\gamma\)
\end{tabbing}
%}
\end{minipage}
}
\end{center}
\caption{Compiling a CNF into d-DNNF. 
Each node \(N\) in the decomposition tree has an associated cache \(\cache_N\);
\(N_l\) and \(N_r\) are the left and right children of node \(N\), respectively;
\(\clause(N)\) returns the clause attached to leaf node \(N\);
\(\atoms(N)\) are the atoms of clauses appearing under node \(N\);
\(\atoms(\alpha)\) returns the atoms appearing in instantiation \(\alpha\); 
\(\project(\alpha,A)\) returns the subset of instantiation \(\alpha\) pertaining to atoms \(A\); and
\(\bdnnfs(\beta)\) returns a d-DNNF of clause \(\beta\).
\label{fig:cnf2ddnnf}}
\end{figure}

Binary decision diagrams (BDDs) are among the most successful representations of propositional
theories \cite{Bryant92}. Two special classes of BDDs, OBDDs and FBDDs, are
especially popular given \cite{FBDDMeinel,FBDDWegener}:
\begin{description}
\item[-] the number of linear-time operations they support and
\item[-] the number of real-world theories that admit efficient OBDD/FBDD representations. 
\end{description}
We now present a linear-time algorithm for converting an FBDD into an equivalent d-DNNF,
showing that d-DNNFs scale as well as FBDDs (which include OBDDs as a subclass).
We start by the formal definitions of BDDs, OBDDs, and FBDDs.
\begin{definition}\label{def:bdd}
A \underline{binary decision diagram (BDD)} over a set of binary variables \(X = \{x_1, \ldots, x_n\}\) is
a directed acyclic graph with one root and at most two leaves labeled \(0\) and \(1\).
Each non-leaf node \(m\) is labeled by a variable \(\var(m) \in X\) and has two outgoing edges
labeled \(0\) and \(1\), where \(\low(m)\) and \(\high(m)\) denote the nodes pointed to
by these edges, respectively. 
The \underline{computation path} for input \((a_1, \ldots, a_n)\), where \(a_i \in \{0,1\}\),
is defined as follows. One starts at the root. At inner node \(m\), where \(\var(m)=x_i\), 
one moves to node \(\low(m)\) if \(a_i = 0\) and to node \(\high(m)\) otherwise. The
BDD represents the Boolean function \(f\) if the computation path for each input 
\((a_1, \ldots, a_n)\) leads to the leaf node labeled with \(f(a_1, \ldots, a_n)\).
\end{definition}
The size of a BDD is measured by the number of nodes it contains.

\begin{definition} 
A binary decision diagram is called a \underline{free BDD (FBDD)} if on each computation path each
variable is tested at most once.
A free BDD is called an \underline{ordered BDD (OBDD)} if on each computation path the variables are tested in
the same order.
\end{definition}
OBDDs are a strict subclass of FBDDs \cite{Bryant91} and have received much consideration in the verification 
literature, where they are used to test the equivalence between the specs of a Boolean function
and its circuit implementation. OBDDs/FBDDs permit such a test to be performed in polynomial time.
Their popularity stems from the existence of efficient OBDD/FBDD representations of many complex,
real-world propositional theories. DNNF is more space-efficient than FBDDs \cite{darwicheD109},
but this should not be surprising as FBDDs admit more linear-time operations than does DNNF. For example,
the equivalence of two DNNFs cannot be decided in polynomial time while it can for FBDDs.

\begin{figure}[tb]
\begin{center}
\fbox{
\begin{minipage}{3.0in}
%{\small
\begin{center}
\underline{{\bf Algorithm~fbdd2ddnnf}}
\end{center}
\begin{tabbing}
/* \(m\) is a node in an FBDD */ \\
{\sc f}\={\sc bdd2ddnnf}\((m)\) \\
 \>i\=f \(\cache(m) \neq \nil\), return \(\cache(m)\) \\
 \>i\=f \(m\) is a leaf node labeled with 1, then \(\gamma \set \true\) \\
 \>\>el\=se if \(m\) is a leaf node labeled with 0, then \(\gamma \set \false\) \\
 \>\>\>else \=\(\gamma \set \fbdd2ddnnf(\low(m))\wedge\neg x_i\) \\
 \>\>\>\>\(\vee \fbdd2ddnnf(\high(m))\wedge x_i\) \\
 \>\>\>\> where \(\var(m) = x_i\) \\
 \>\(\cache(m) \set \gamma\) \\
 \>return \(\gamma\)
\end{tabbing}
%}
\end{minipage}
}
\end{center}
\caption{Converting an FBDD into a d-DNNF. 
\(\cache(m)\) stores the d-DNNF computed for the FBDD rooted at node \(m\).
\label{fig:fbdd2ddnnf}}
\end{figure}

\begin{figure*}[tb]
\begin{center}
\mbox{\epsfxsize=105mm \leavevmode \epsffile{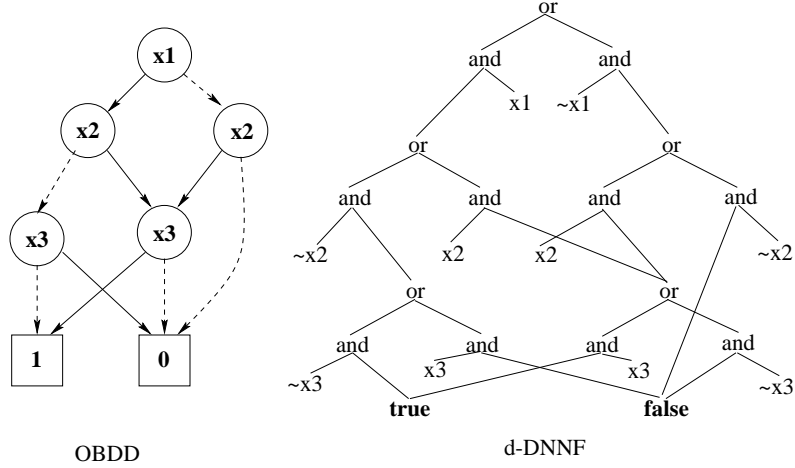}}
\end{center}
\caption{An FBDD and its corresponding DNNF, which are equivalent to
\((x_2\wedge x_3) \vee (x_1\wedge\neg x_2\wedge \neg x_3)\). 
\label{fig:bdd-dnnf}}
\end{figure*}

Figure~\ref{fig:fbdd2ddnnf} depicts a recursive algorithm for converting an FBDD into a DNNF, showing
that DNNFs scale as well as FBDDs. The algorithm should be called on the root of given FBDD:
\begin{theorem}\cite{darwicheD109}
The algorithm of Figure~\ref{fig:fbdd2ddnnf} takes time linear in the size of given FBDD and returns
a DNNF of the theory represented by the given FBDD.
\end{theorem}
\begin{theorem}
The DNNF returned by the algorithm of Figure~\ref{fig:fbdd2ddnnf} is deterministic.
\end{theorem}
This has major implications on our reported results regarding truth maintenance and belief revision 
systems, as it proves, constructively, that we can build efficient truth maintenance and belief 
revision systems for any propositional theory which has an efficient FBDD representation.
Figure~\ref{fig:bdd-dnnf} depicts an FBDD and its corresponding d-DNNF as generated by the algorithm
of Figure~\ref{fig:fbdd2ddnnf}.

\section{Counting Models of d-DNNF} \label{sec:counting}

We now turn to an operation on d-DNNF which is of major significance to a number of AI
applications, including belief revision, truth maintenance and diagnosis. 
Specifically, given a d-DNNF \(\Delta\) 
and a set of literals \(\S\), we describe two traversal operations each taking
linear time. By the end of the first traversal, we will be able to count the models of
\(\Delta \cup \S\). By the end of the second traversal, we will be able to count the models of:
\begin{enumerate}
\item \(\Delta \cup \S \cup \{l\}\) for every literal \(l \not \in \S\);
\item \(\Delta \cup \S \setminus\{l\}\) for every literal \(l \in \S\);
\item \(\Delta \cup \S \setminus\{l\}\cup\{\neg l\}\) for every literal \(l \in \S\).
\end{enumerate}
{\em That is, once we traverse the d-DNNF twice, we will be able to obtain each of these
counts using constant-time, lookup operations.}
As we shall see in the following section, these counts are all we need to implement
an interesting number of AI applications.

The traversal will not take place on the d-DNNF itself, but on a secondary structure that we call
the {\em counting graph.} Without loss of generality, we will assume from here on that the d-DNNF is smooth:
\begin{definition}
A DNNF is \underline{smooth} iff
\begin{enumerate}
\item every literal and its negation appear in the DNNF;
\item for any disjunction \(\bigvee_i \alpha_i\) in the DNNF, we have 
\(\atoms(\bigvee_i \alpha_i)=\atoms(\alpha_i)\) for every \(\alpha_i\). 
\end{enumerate}
\end{definition}
The d-DNNF in Figure~\ref{fig:odd} is smooth as it satisfies these two conditions. We can easily
make a DNNF smooth using two operations:
\begin{enumerate}
\item If the negation of literal \(l\) does not appear in the DNNF, replace the occurrence of 
\(l\) with \(l \vee (\neg l \wedge \false)\);
\item For each disjunction \(\bigvee_i \alpha_i\), replace the disjunct \(\alpha_i\) with
\(\alpha_i \wedge \bigwedge_{A \in \Sigma} (A \vee \neg A)\), where \(\Sigma\) are the atoms
appearing in \(\bigvee_i \alpha_i\) but not in \(\alpha_i\).
\end{enumerate}
These operations preserve both the decomposability and determinism of a DNNF.
They may increase the size of given DNNF but only by a factor of \(O(n)\), where
\(n\) is the number of atoms in the DNNF. This increase is quite minimal in practice though.
Note that the d-DNNFs generated by the algorithm of Figure~\ref{fig:fbdd2ddnnf} satisfy the 
first condition; and the d-DNNFs generated by the algorithm of Figure~\ref{fig:cnf2ddnnf} satisfy the 
second condition as long as \(\bdnnfs(\clause(N)\mid\alpha)\) satisfies some simple conditions;
see \cite{darwicheD113}.

\begin{figure*}[tb]
\begin{center}
\mbox{\epsfxsize=105mm \leavevmode \epsffile{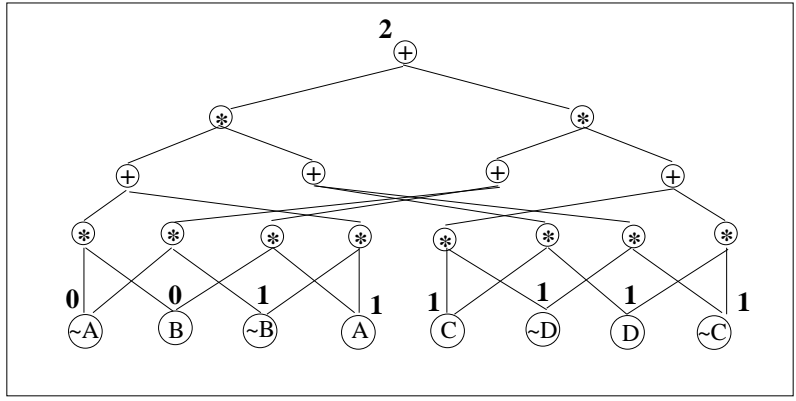}}
\end{center}
\caption{A counting graph of the DNNF \(\Delta\) in Figure~\ref{fig:odd} evaluated under \(\S = A,\neg B\).
The evaluation indicates that \(\Delta \cup \S\) has two models (\(A\wedge\neg B\wedge C\wedge D\)
and \(A\wedge\neg B\wedge \neg C\wedge \neg D\) in this case).
\label{fig:odd-cgraph}}
\end{figure*}

The counting graph of a d-DNNF is a function of many variables represented as a rooted DAG. 
\begin{definition}
The \underline{counting graph} of a smooth d-DNNF is a labeled, rooted DAG. It contains a node 
labeled with \(l\) for each literal \(l\), a node labeled with \(+\) for each or-node, and a 
node labeled with \(*\) for each and-node in the d-DNNF. There is an edge between two nodes in
the counting graph iff there is an edge between their corresponding nodes in the d-DNNF.
\end{definition}
Figure~\ref{fig:odd-cgraph} depicts the counting graph of the d-DNNF in Figure~\ref{fig:odd}.
The size of a counting graph is therefore equal to the size of its corresponding d-DNNF. 
We will see now how such a graph can be used to perform the counting operations we are interested in.

\begin{definition} \label{def:value}
The \underline{value} of a node \(N\) in a counting graph under a set of literals \(\S\) is defined 
as follows:
\begin{itemize}
\item[-] \(\val(N) = 0\) if \(N\) is labeled with literal \(l\) and \(\neg l \in \S\);
\item[-] \(\val(N) = 1\) if \(N\) is labeled with literal \(l\) and \(\neg l \not \in \S\);
\item[-] \(\val(N) = \prod_i \val(N_i)\) if \(N\) is labeled with \(*\), where \(N_i\) are the 
children of \(N\);
\item[-] \(\val(N) = \sum_i \val(N_i)\) if \(N\) is labeled with \(+\), where \(N_i\) 
are the children of \(N\).
\end{itemize}
The value of a counting graph \(G\) under literals \(\S\), written \(G(\S)\), 
is the value of its root under \(\S\).
\end{definition}

Here's our first counting result.
\begin{theorem} \label{theo:count}
Let \(\Delta\) be a smooth d-DNNF, \(\S\) be a set of literals, and let \(G\) be the
counting graph of \(\Delta\). The value of \(G\) under \(\S\) is the number of models of \(\Delta \cup \S\):
\[ G(\S) = \mc(\Delta \cup \S). \]
\end{theorem}
Note that \(G(\S) > 0\) iff \(\Delta \cup \S\) is consistent. Therefore, by traversing the counting
graph once we can test the consistency of d-DNNF \(\Delta\) conjoined with any set of literals \(\S\). 
Figure~\ref{fig:odd-cgraph} depicts the counting graph of d-DNNF \(\Delta\) in Figure~\ref{fig:odd},
evaluated under the literals \(\S = A,\neg B\). 
This indicates that \(\Delta \cup \{A,\neg B\}\) has two models.

\begin{figure*}[tb]
\begin{center}
\mbox{\epsfxsize=105mm \leavevmode \epsffile{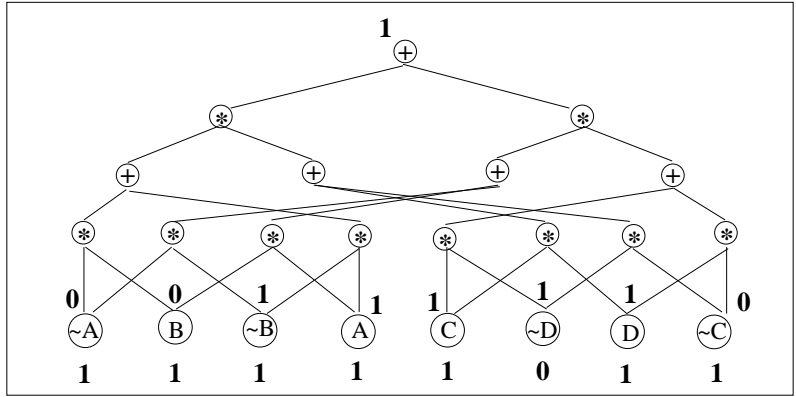}}
\end{center}
\caption{A counting graph of the DNNF \(\Delta\) in Figure~\ref{fig:odd} evaluated under 
literals \(\S = A,\neg B,C\), indicating one model of \(\Delta \cup \S\) 
(\(A \wedge \neg B \wedge C \wedge D\) in this case).
Partial derivatives are shown below the leaves.
\label{fig:odd-cgraph-pd}}
\end{figure*}

We now present the central result in this paper.
First, we note that when viewing a counting graph \(G\) as a function of many variables, 
we will use \(V_l\) to denote the variable (node) which corresponds to literal \(l\).
Second, we can talk about the partial derivative of \(G(\S)\) with respect to any of 
its variables \(V_l\), \(\partial G(\S) / \partial V_l\). 
Due to the decomposability of d-DNNF, the function \(G(\S)\) is linear in each of its 
variables. Therefore, the change to the count \(G(\S)\) as a result of adding, removing
or flipping a literal in \(\S\) can be obtained from the partial derivatives, without
having to re-evaluate the counting graph \(G\). This leads to the following consequential
result:
\begin{theorem} \label{theo:pds}
Let \(\Delta\) be a smooth d-DNNF, \(\S\) be a set of literals, and let \(G\) be the
counting graph of \(\Delta\). We have:
\begin{description}
\item [Assertion:] When \(l,\neg l \not \in \S\):
\[ \mc(\Delta \cup \S \cup \{l\}) = \partial G(\S) / \partial V_l.\]

\item [Retraction:] When \(l \in \S\):
\[ \mc(\Delta \cup \S \setminus \{l\}) = 
\partial G(\S) / \partial V_l + \partial G(\S) / \partial V_{\neg l}.\]

\item [Flipping:] When \(l \in \S\):
\begin{eqnarray*}
\lefteqn{\mc(\Delta \cup \S \setminus \{l\} \cup \{\neg l\})}  \\
& = & G(\S) - \partial G(\S) / \partial V_l + \partial G(\S) / \partial V_{\neg l}. 
\end{eqnarray*}
\end{description}
\end{theorem}
Therefore, if we can compute the partial derivative of \(G(\S)\) with respect to each of its variables,
then we can count the models of \(\Delta \cup \S\) under the assertion of new literals not in \(\S\), and
under the retraction or flipping of literals in \(\S\). Figure~\ref{fig:odd-cgraph-pd} depicts the value 
of each of these partial derivatives for the d-DNNF in Figure~\ref{fig:odd}. The counting graph is
evaluated under literals \(\S = A,\neg B, C\) and the partial derivatives are shown below each variable.
According to these derivatives, we have:
\begin{description}
\item [Assertion:]
\(\mc(\Delta \cup \S \cup \{D\}) = 1\) and \(\mc(\Delta \cup \S \cup \{\neg D\}) = 0\).
This immediately tells us that \(\Delta \cup \S \models D\).

\item [Retraction:]
\(\mc(\Delta \cup \S \setminus \{A\}) = 1+1=2\); 
\(\mc(\Delta \cup \S \setminus \{\neg B\}) = 1+1=2\); and
\(\mc(\Delta \cup \S \setminus \{C\}) = 1+1=2\).
Therefore, retracting any literal in \(\S\) increases the number of models to \(2\).

\item [Flipping:]
\(\mc(\Delta \cup \S \setminus \{A\} \cup \{\neg A\}) = 1-1+1=1\);
\(\mc(\Delta \cup \S \setminus \{\neg B\} \cup \{B\}) = 1-1+1=1\);
\(\mc(\Delta \cup \S \setminus \{C\} \cup \{\neg C\}) = 1-1+1=1\).
Therefore, flipping any literal in \(\S\) will not change the number of models 
(although it does change the model itself).
\end{description}

There is one missing link now: How do we compute the partial derivatives of a counting graph with
respect to each of its variables? This actually turns out to be easy due to results in \cite{iri84,Graz90}
which show how to evaluate and simultaneously compute all partial derivatives of a function by simply
traversing its {\em computation graph} in linear time. Although \cite{Graz90} casts such computation 
in terms of summing weights of paths in such a graph, we present a more direct implementation here.
In particular, if we let \(\pd(N)\) denote the partial derivative of \(G(\S)\) with respect to a 
node \(N\) in the counting graph, then \(\pd(N)\) is the summation of contributions made by parents 
\(M\) of \(N\):
\begin{eqnarray*}
\pd(N) & = & \left\{
\begin{array}{ll}
1, & \mbox{N is the root node;} \\
\displaystyle \sum_{M} \cpd(M,N), & \mbox{otherwise;} 
\end{array}
\right.
\end{eqnarray*}
where the contribution of parent \(M\) to its child \(N\) is computed as follows:
\[ \cpd(M,N) =
\left\{
\begin{array}{ll}
\pd(M), & \mbox{\(M\) is a \(+\)node;} \\
\displaystyle \pd(M)\prod_{K \neq N} \val(K), & \mbox{\(M\) is a \(\star\)node;} 
\end{array}
\right.
\]
where \(K\) is a child of \(M\). This computation can be performed by first traversing the
counting graph once to evaluate it, assigning \(\val\) to each node \(N\), and then 
traversing it a second time, assigning \(\pd\) for each node \(N\). We are then mainly interested in
\(\val(N)\) where \(N\) is the root node, and \(\pd(N)\) where \(N\) is a leaf node.

Therefore, both the value of a counting graph under some literals \(\S\) and the
values of each of its partial derivatives under \(\S\) can be computed by traversing the
graph twice. Once to compute the values, and another to compute the partial derivatives.
Note that such traversal needs to be redone once the set of literals \(\S\) changes.

We close this section by pointing out that partial differentiation turns out to play
a key role in probabilistic reasoning as well. Specifically, we present a comprehensive
framework for probabilistic reasoning in \cite{darwicheNM00-DIF} based on compiling a
Bayesian network into a polynomial and then reducing a large number of probabilistic
queries into the computation of partial derivatives of the compiled polynomial.

\section{Complete, Linear-Time Truth Maintenance} \label{sec:tms}

We now turn to some applications of the results in the previous section. That is, what
can we do once we are able to count models under the conditions stated above?

We first consider truth maintenance systems and show how our results allow us
to implement complete truth maintenance systems which take linear time on 
two important classes of propositional theories: those with bounded treewidth,
and those admitting a linear FBDD representation. For each class of such theories,
we can compile a smooth d-DNNF \(\Delta\) in linear time and then use it for truth maintenance
as follows.\footnote{We are assuming that smoothing a d-DNNF does not increase its
size by more than a constant factor.}

A truth maintenance system takes a set of clauses \(\Gamma\) and a set of literals
\(\S\) and tries to determine for each literal \(l\) whether \(\Gamma \cup \S \models l\).
The most common truth maintenance system is the one based on closing 
\(\Gamma \cup \S\) under unit resolution \cite{dekleerBOOK93}. Such a system takes linear time, but is
\underline{incomplete.} Given that the set of literals in \(\S\) changes to \(\S^\prime\), 
the goal of a truth maintenance system is then to update the truth of each
literal under the new ``context'' \(\S^\prime\). Sometimes, clauses in \(\Gamma\) are
retracted and/or asserted. A truth maintenance system is expected to update
the truth of literals under such clausal changes too.

Our model-counting results allow us to implement a \underline{complete} 
truth maintenance system as follows. We compile the theory \(\Gamma\) into a smooth d-DNNF
\(\Delta\) and construct the counting graph \(G\) of \(\Delta\). Given any set
of literal \(\S\), we evaluate \(G\) under \(\S\) and compute its partial derivatives
also under \(\S\). This can be done in time linear in the size of \(\Delta\).
We are now ready to answer all queries expected from a truth maintenance system
by simple, constant-time, look-up operations:
\begin{description}
\item Literal \(l\) is entailed by \(\Delta \cup \S\) iff \(\Delta \cup \S \cup \{\neg l\}\)
has no models: \(\partial G(\S) / \partial V_{\neg l} = 0\).

\item Retracting literal \(l\) from \(\S\) will render \(\Delta \cup \S\) consistent
iff \(\Delta \cup \S \setminus \{l\}\) has at least one model:
\(\partial G(\S) / \partial V_{l} + \partial G(\S) / \partial V_{\neg l} > 0\).

\item Flipping literal \(l\) in \(\S\) will render \(\Delta \cup \S\) consistent
iff \(\Delta \cup \S \setminus \{l\} \cup \{\neg l\}\) has at least one model:
\(G(\S) - \partial G(\S) / \partial V_{l} + \partial G(\S) / \partial V_{\neg l} > 0\).
\footnote{Note that the flipping of literals is outside the scope of classical truth 
maintenance systems in the sense that they must retract \(l\) and then assert \(\neg l\), 
taking linear time, to perform the above operation.}
\end{description}

If we want to reason about the assertion/retraction of clauses in theory \(\Gamma\),
we can replace each clause \(\alpha\) in \(\Gamma\) by \(C_\alpha \equiv \alpha\),
where \(C_\alpha\) is a new atom that represents the truth of clause \(\alpha\). We then
compile the extended theory \(\Gamma\) into d-DNNF \(\Delta\).
To assert all clauses 
initially, we have to include all atoms \(C_\alpha\) in the set of literals \(\S\).
The assertion/retraction of clauses can then be emulated by the assertion/retraction
of atoms \(C_\alpha\). For example, in case of a contradiction, we can ask whether 
removing a clause \(\alpha\) will resolve the contradiction by asking whether 
\(\Delta \cup \S \setminus \{C_\alpha\}\) has more than one model:
\[ \partial G(\S) / \partial V_{C_\alpha} + \partial G(\S) / \partial V_{\neg C_\alpha} > 0. \].

\section{Complete, Linear-Time Belief Revision} \label{sec:br}

We now turn to a second major application of model counting on d-DNNF: the implementation
of a very common class of belief revision systems, which is adopted in model-based diagnosis
and in certain forms of default reasoning. The problem here is as follows. We have a set of special atoms
\(\D = \{d_1,\ldots,d_n\}\) in the theory \(\Gamma\) which represent defaults. Typically, we 
assume that all of these defaults are true, allowing us to draw some default conclusions. 
We then receive some evidence \(\S\) (a set of literals) which is inconsistent
with \(\Gamma \cup \D\). We therefore know that not all defaults are true and some must be
retracted---that is, some \(d_i\)s will have to be replaced by \(\neg d_i\) in \(\D\). 
Our goal then is to identify a set of literals \(\D^\prime\) such that
\begin{enumerate}
\item \(d_i \in \D^\prime\) or \(\neg d_i \in \D^\prime\) for all \(i\);
\item \(\Gamma \cup \D^\prime \cup \S\) is consistent;
\item the number of negative literals in \(\D^\prime\) is minimized;
\end{enumerate} 
and then then report the truth of every literal under the new set of defaults \(\D^\prime\). 
Note that there may be more than one set of defaults \(\D^\prime\) that satisfies the above 
properties. In such a case, a literal holds after the revision process only if it holds under 
\(\Gamma \cup \D^\prime \cup \S\) for every \(\D^\prime\).

How can we do this? As we shall see now, if \(\Gamma\) is a smooth d-DNNF, then all 
of this can be done in time linear in the size of \(\Gamma\)!

\begin{figure*}[tb]
\begin{center}
\mbox{\epsfxsize=105mm \leavevmode \epsffile{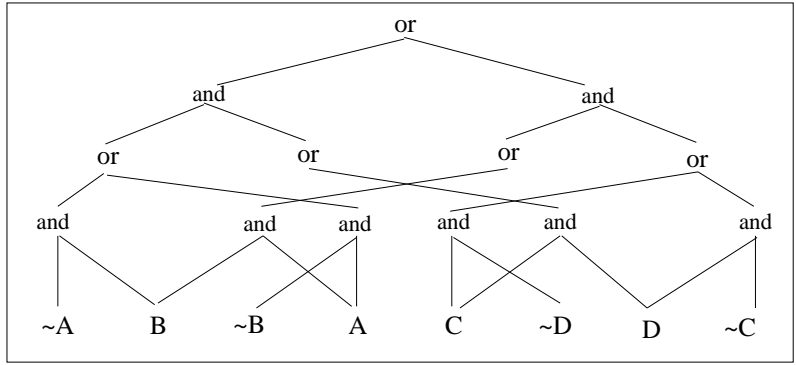}}
\end{center}
\caption{A minimized d-DNNF.
\label{fig:odd-min2}}
\end{figure*}

This works exactly as in the previous section, except that we have to {\em minimize} the d-DNNF
first, a process which eliminates some of the d-DNNF models \cite{darwicheD109}. To define this
minimization process more precisely, we need the following definition first:
\begin{definition}
If \(\Sigma\) is a set of atoms, then the \(\Sigma\)-cardinality of a model is the number of
atoms in \(\Sigma\) that the model sets to false.
\end{definition}
To \(\Sigma\)-minimize a theory \(\Gamma\) is to convert it into another theory whose models are 
exactly the models of \(\Gamma\) having a minimum \(\Sigma\)-cardinality.
Consider the d-DNNF \(\Gamma\) in Figure~\ref{fig:odd} for an example and suppose that 
\(\Sigma = \{A,B,C,D\}\); that is, we want to minimize the d-DNNF with respect to each of its atoms. 
This theory has eight models, each having an odd cardinality (one or three). If we \(\Sigma\)-minimize this 
d-DNNF, we obtain another with four models, shown in Figure~\ref{fig:odd-min2}. 

Given these definitions, we can re-phrase the problem of belief revision (stated above) as follows.
Let \(\Sigma\) be a set of atoms representing defaults, and let \(\Gamma\) be a smooth d-DNNF.
Given observation \(\S\), \(\Sigma\)-minimize the theory \(\Gamma \cup \S\) to yield 
\(\Delta\) and then report on the truth of each literal under the minimized theory 
\(\Delta\).

\begin{figure*}[tb]
\begin{center}
\mbox{\epsfxsize=105mm \leavevmode \epsffile{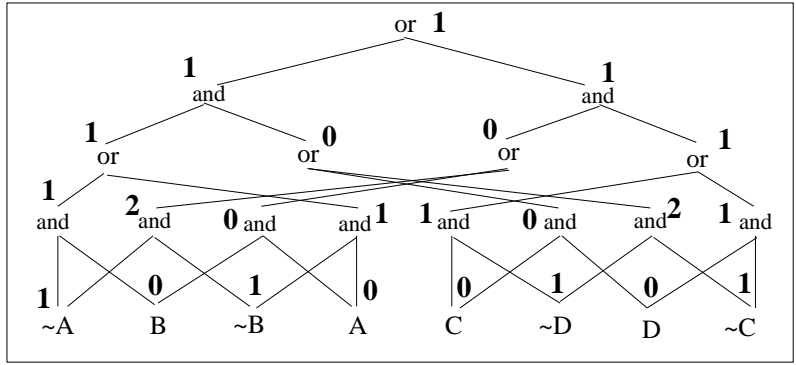}}
\end{center}
\caption{Assigning a cardinality to each node in a d-DNNF with \(\Sigma = \{A,B,C,D\}\). 
\label{fig:odd-min}}
\end{figure*}

As it turns out, one can minimize a smooth d-DNNF in linear time, to yield another smooth 
d-DNNF to which we can apply the techniques of the previous section.
We now describe the process of minimizing a DNNF which is described in more details in \cite{darwicheD109}. 
We do this in a two-step process:
\begin{enumerate}
\item We assign a cardinality to every node in the d-DNNF as follows:
\begin{enumerate}
\item each literal whose atom is not in \(\Sigma\) gets cardinality zero;
\item each positive literal whose atom is in \(\Sigma\) gets cardinality zero; 
\item each negative literals whose atom is in \(\Sigma\) gets cardinality one; 
\item the cardinality of a disjunction is the min of its disjuncts' cardinalities; 
\item the cardinality of a conjunction is the summation of its conjuncts' cardinalities.
\end{enumerate}

\item For each or-node \(N\) and its child \(M\), we delete the edge connecting \(N\) and \(M\)
if the cardinality of \(N\) is smaller than the cardinality of \(M\).
\end{enumerate}
Figure~\ref{fig:odd-min} depicts the result of assigning cardinalities to the d-DNNF of Figure~\ref{fig:odd}, 
and Figure~\ref{fig:odd-min2} depicts the result of deleting some of its edges. 
This is the minimized d-DNNF and it has four models:
\begin{description}
\item \(\neg A\wedge B \wedge C \wedge D\); 
\item \(A\wedge \neg B \wedge C \wedge D\); 
\item \(A\wedge B \wedge \neg C \wedge D\); 
\item \(A\wedge B \wedge C \wedge \neg D\).
\end{description}
Once we have minimized the d-DNNF, we can apply the results of the previous section to obtain
the answers we want. 

\begin{figure*}[tb]
\begin{center}
\mbox{\epsfxsize=105mm \leavevmode \epsffile{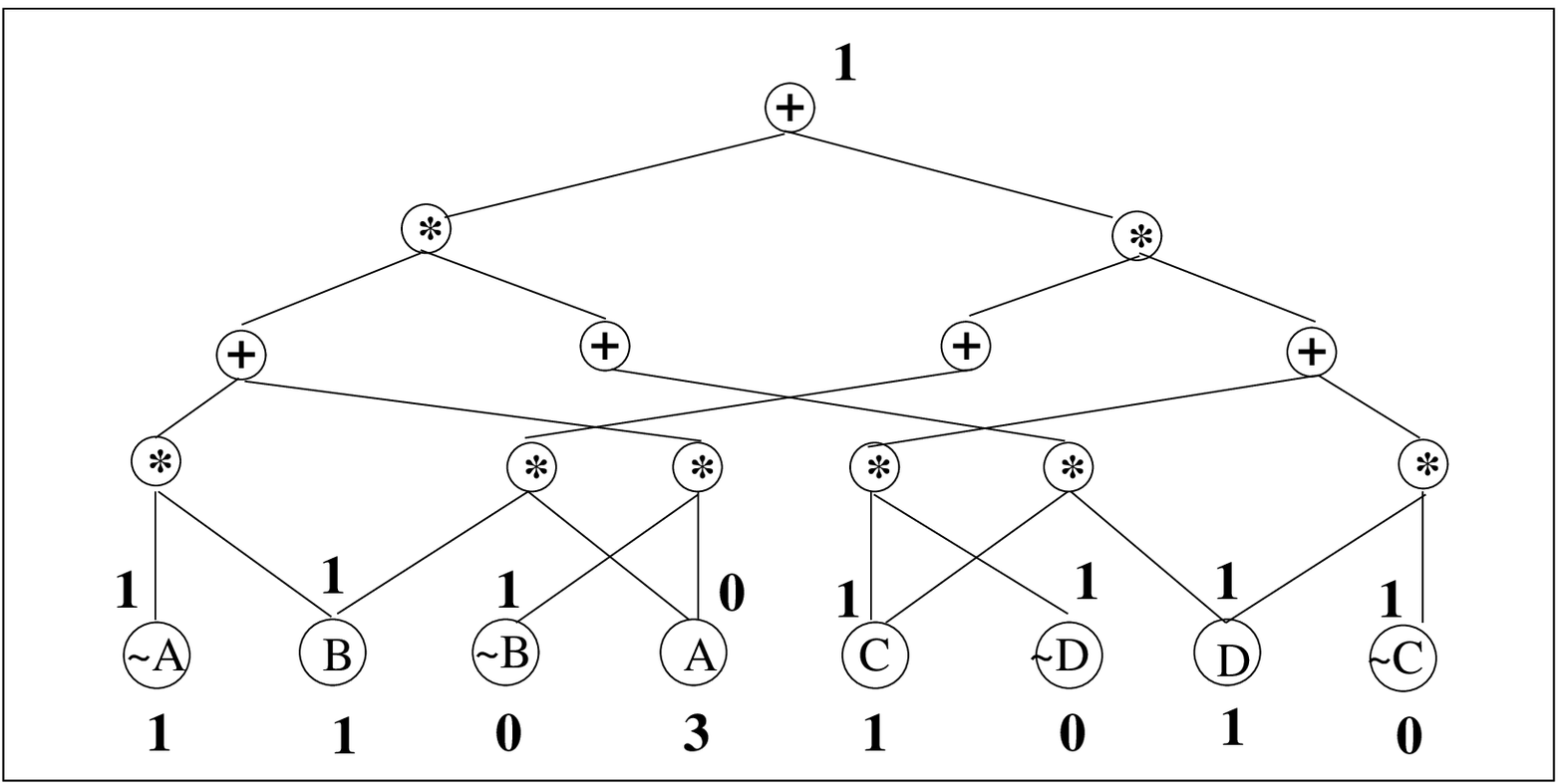}}
\end{center}
\caption{The counting graph of the minimized d-DNNF in Figure~\ref{fig:odd-min2}
evaluated under \(\S = \{\neg A\}\). Partial derivatives are shown below leaves.
\label{fig:odd-min-cgraph}}
\end{figure*}

As an example, Figure~\ref{fig:odd-min-cgraph} depicts the counting graph of the 
minimized d-DNNF \(\Delta\) in Figure~\ref{fig:odd-min2}, with its value and partial derivatives
computed under the observation \(\S = \{\neg A\}\). From these partial derivatives
and Theorem~\ref{theo:pds}, we immediately get:
\begin{description}
\item [Assertion:] \(\mc(\Delta \cup \S \cup \{\neg B\}) = 0\);
\(\mc(\Delta \cup \S \cup \{\neg C\}) = 0\);
\(\mc(\Delta \cup \S \cup \{\neg D\}) = 0\).
That is, \(B, C\) and \(D\) are all entailed by \(\Delta \cup \S\).

\item [Retraction:] \(\mc(\Delta \cup \S \setminus \{\neg A\}) = 1+3=4\).
That is, we have four models if we retract \(\neg A\).

\item [Flipping:] \(\mc(\Delta \cup \S \setminus \{\neg A\} \cup \{A\}) = 1-1+3=3\).
That is, we have three models if we flip \(\neg A\) to \(A\).
\end{description}

\section{Predicting the Behavior of Broken Devices}
The above results have direct application to model-based diagnosis, where
\(\Delta\) is the device description, \(\S\) is the device observation and
\(\D\) contains the health modes \(\ok_1,\ldots,\ok_n\). Initially, we assume
that all device components are working normally, but then find some observation
\(\S\) such that \(\Delta \cup \D=\{\ok_1,\ldots,\ok_n\} \cup \S\) is inconsistent.

To regain consistency we must postulate that some of the components are not
healthy, therefore, flipping some of the \(\ok_i\)s into \(\neg \ok_i\) in the
set \(\D\). Assuming a smallest number of faults, we want to minimize the number 
of unhealthy components needed to regain consistency. A set \(\D^\prime\) such that:
\begin{enumerate}
\item \(\ok_i \in \D^\prime\) or \(\neg \ok_i \in \D^\prime\) for all \(i\);
\item \(\Delta \cup \D^\prime \cup \S\) is consistent;
\item the number of negative literals in \(\D^\prime\) is minimized;
\end{enumerate}
is called a {\em minimum-cardinality diagnosis} and one goal of model-based diagnosis
to enumerate such diagnoses \cite{darwicheJAIR-SSD,dekleer:characterizing-diagnoses}. 

Another practical problem, however, which has received much less attention in model-based
diagnosis is the following: Assuming a smallest number of faults, what is the truth value of 
every literal appearing in the device description \(\Delta\)? That is, we do not want to
know what the minimum-cardinality diagnoses are. Instead --- and under the assumption
that one of them has materialized --- we want to predict the behavior of the given
faulty device. But this is exactly the problem we have treated in the previous section.

Therefore, our results allow us to predict the value of each device port (literal \(l\)) 
in a broken device, assuming that the number of broken components is minimal, in time 
linear in the size of device description \(\Delta\), as long as \(\Delta\) is a smooth
d-DNNF. We are unaware of any similar complexity result for model-based diagnosis.

\section{Conclusion} \label{sec:conclusion}
We have identified two classes of propositional theories, those which have a bounded treewidth,
and those which have a linear-sized FBDD representation. We have shown that each of these
classes of theories can be converted in linear time into a tractable form that we called
deterministic DNNF, d-DNNF. We have also defined linear-time, model-counting operations on 
d-DNNF, allowing us to implement (a) linear-time, complete truth maintenance systems and 
(b) linear-time, complete belief revision systems for the two identified classes of 
propositional theories. Our results also have major implications on the practice of model-based
diagnosis as they allow us to efficiently predict the behavior of a broken device,
assuming a smallest number of broken components.

\end{document}